\documentclass{article}

\usepackage{microtype}
\usepackage{graphicx}
\usepackage{subfigure}
\usepackage{booktabs} 

\usepackage{hyperref}

\usepackage{localmacros}

\usepackage[accepted]{icml2024}

\usepackage{amsmath}
\usepackage{amssymb}
\usepackage{mathtools}
\usepackage{amsthm}

\graphicspath{{figures/}}

\theoremstyle{plain}

\theoremstyle{definition}

\theoremstyle{remark}

\numberwithin{equation}{section}

\usepackage[textsize=tiny]{todonotes}

\icmltitlerunning{Neural Incremental Data Assimilation}

\begin{document}

\twocolumn[
\icmltitle{Neural Incremental Data Assimilation}




\begin{icmlauthorlist}
\icmlauthor{Matthieu Blanke}{argo}
\icmlauthor{Ronan Fablet}{imt}
\icmlauthor{Marc Lelarge}{argo}
\end{icmlauthorlist}

\icmlaffiliation{argo}{Inria Paris, DI ENS, PSL Research University}
\icmlaffiliation{imt}{IMT Atlantique, Brest, France}

\icmlcorrespondingauthor{Matthieu Blanke}{matthieu.blanke@inria.fr}

\icmlkeywords{Deep learning, data assimilation, 4D-Var, diffusion}

\vskip 0.3in
]



\printAffiliationsAndNotice{}  

\begin{abstract}
    Data assimilation is a central problem in many geophysical applications, such as weather forecasting. It aims to estimate the state of a potentially large system, such as the atmosphere, from sparse observations, supplemented by prior physical knowledge. The size of the systems involved and the complexity of the underlying physical equations make it a challenging task from a computational point of view. Neural networks represent a promising method of emulating the physics at low cost, and therefore have the potential to considerably improve and accelerate data assimilation. In this work, we introduce a deep learning approach where the physical system is modeled as a sequence of coarse-to-fine Gaussian prior distributions parametrized by a neural network. This allows us to define an assimilation operator, which is trained in an end-to-end fashion to minimize the reconstruction error on a dataset with different observation processes. We illustrate our approach on chaotic dynamical physical systems with sparse observations, and compare it to traditional variational data assimilation methods.
\end{abstract}


\section{Introduction}
\label{section:assimilation_introduction}

Artificial intelligence is transforming many fields, and has a growing number of applications in industry. In the sciences, it has the potential to considerably accelerate the scientific process. Geophysics and weather forecasting are areas where deep learning is particularly active, with recent months seeing an explosion in the number of large neural models for the weather forecasting problem~\citep{pathak2022fourcastnet, lam2022graphcast, hoyer2023neural}, building on reanalysis datasets such as~ERA5~\citep{munoz2021era5} for training. In this work, we focus on the data assimilation problem that underpins weather forecasting: tomorrow's weather forecast is based on today's weather conditions, which are not directly measured, but are estimated from few observations. Data assimilation is the inverse problem of estimating the geophysical state of the globe on the basis of these sparse observations and of prior knowledge of the physics. The estimated state then serves as the starting point for forecasting.
While deep learning models are revolutionizing the forecasting problem, they have yet to be applied operationally to data assimilation.

The application of neural networks to inverse problems is an active area of research. The general idea consists in training a neural network to reconstruct a signal, using for training examples a dataset of simulated physical states serving as ground truth. For the data assimilation problem, several approaches have been proposed to incorporate a deep learning in the loop.~\cite{arcucci2021deep} propose a sequential scheme where a neural network is trained at regular time steps to combine data assimilation and the forecasting model. Recently, the success of diffusion models for imaging~\citep{ho2020denoising} has led to the development of so-called "plug and play" methods, where the neural network is trained to learn a prior~\citep{laumont2022bayesian}. Once trained, the neural prior can be used to solve a large number of inverse problems. In this line of work,~\cite{NEURIPS2023_7f7fa581} proposed a data assimilation method based on a diffusion model. Another type of approaches called ``end-to-end'' aim at directly training a neural network to minimize the reconstruction error. They have the benefit of training the network directly on the task of interest, but the versatility of the trained model with respect to the different observational processes is challenging. An end-to-end neural reconstruction algorithm is proposed in~\citep{fablet2021learning}, and aims at learning the prior distribution of the signal by defining the reconstruction as a maximum a posterior estimate, leading to a bi-level optimization problem. However, the complex prior induced by the neural network may hamper the convergence of this estimate, as it relies on non-convex optimization. Instead, we explore a model where the prior has a sufficiently simple structure to guarantee a convex posterior distribution.

\paragraph{Contributions} In this work, we present a neural method for data assimilation. We introduce a data assimilation operator parametrized by a neural~Gaussian prior, that is designed to locally improve the likelihood of an estimate. Our model is trained to minimize the reconstruction error in an end-to-end fashion. We show how this operator may be iterated to reconstruct complex signals. The effectiveness of our method is demonstrated on simulated nonlinear physical systems.
We also show how our method may be used to enhance traditional data assimilation methods.

\section{The data assimilation inverse problem}
\label{section:assimilation_background}

The aim of data assimilation is to reconstruct a state~$x\in \R^d$ from partial noisy measurements~${y \in \R^m}$ of that state~\citep{bouttier2002data, bocquet2014introduction}.
For meteorological applications, for instance, the state~$x$ represents the physical quantities on a grid representing the globe, and the observations~$y$ are partial measurements, from different sources: in situ measurements, weather balloons, satellites,~\textit{etc}. These measurements may be very sparse, with an observation rate~$m/d$ that may be of the order~$1\%$, so we cannot generally hope to recover the state as a function of the observations alone. Indeed, for a given observation vector~$y$, a large number of states are compatible, making data assimilation an inverse problem. To reconstruct the state, we need to supplement the partial observations with another source of prior information on the state, which comes from our physical or statistical knowledge of the problem. 

\begin{sloppypar}
    The data assimilation problem is then as follows. Given partial observations~$y$ and prior information on the state, the aim is to estimate the most probable underlying state~$x$. The Bayesian probabilistic framework lends itself well to the mathematical formalization of the problem : the theoretical information about the state physics is captured by a prior distribution~$x \sim p(x)$, and the noisy, partial observations of~$x$ can be modeled as~$y|x \sim h(x) + \xi$, with~$h$ the observation process, and an unbiased additive noise that is typically assumed to be Gaussian~$\xi \sim \mathcal{N}(0, R)$ and independent of~$x$. Then, data assimilation can be seen as the estimation of the state maximizing the state posterior distribution~${p(x|y)= p(x)p(y|x)/p(y)}$. Under the assumption of Gaussian observational noise, this can be formulated as the following minimization problem
    
    
    \begin{equation}
        \label{problem:MAP_interpolation}
        \underset{x \in \R^d}{\mathrm{minimize}} \quad U(x) + \frac{1}{2} \Vert h(x) - y \Vert^2_{R^{-1}},
    \end{equation}
    with~$U(x) = -\log p(x)$, and where we have adopted the notation~${\Vert z \Vert_C = \sqrt{\transp{z}Cz}}$ for a positive definite matrix~$C$. We assume for simplicity that the observation function~$h$ is known, although it may be only partially known in some cases, such as remote sensing~\citep{liang2005quantitative} or medical imaging~\citep{rangayyan2024biomedical}.
\end{sloppypar}

\paragraph{Problem size}
For weather prediction, the state~$x$ represents the geophysical variables on a large spatial grid. It is hence a signal of very high dimension with typically~$d\sim 10^6$ or even $d\sim 10^9$.  The size of the data assimilation problem makes the computations and memory costs very heavy, severely limiting the computational budget of any numerical method. In the development of new learning-based methods, it is essential to keep this computational constraint in mind if we hope to scale up to real-size systems.


\subsection{Least-squares Gaussian interpolation}

The first approach considered for data assimilation is naturally that of a linear-quadratic model. Assuming a Gaussian a priori on the state~$x \sim \mathcal{N}(\mu, P)$ 
and a linear observation function~$h(x) = Hx$, with~$H \in \R^{m \times d}$, the variational Bayesian formulation for data assimilation~\eqref{problem:MAP_interpolation} becomes a quadratic least-squares problem:
\begin{equation}
    \label{problem:gaussian_interpolation}
   \underset{x \in \R^d}{\mathrm{minimize}} \quad 
   \frac{1}{2}  \Vert x - \mu \Vert^2_{P^{-1}}
   +
  \frac{1}{2}  \Vert Hx - y\Vert^2_{R^{-1}}
\end{equation}
whose maximum a posteriori solution takes the form 
\begin{equation}
    \label{eq:linear_interpolator}
    x_{\mathrm{MAP}}(y; \mu, P) := \mu + K(y-H\mu),
\end{equation}
with the $H$-dependent Kalman gain
\begin{equation}
    \label{eq:kalman_obs}
    K = P \transp{H} (H P \transp{H} + R)^{-1} \; \in \R^{m \times d}.
\end{equation}
In the remainder of this work, the dependence with respect to~$H$ is implicitly assumed in all quantities that depend on the observation vector~$y$.


For meteorological applications, the state~$x$ that is optimized for is a snapshot of the set of geophysical variables at a given time, when the observations have been collected. The background term~$\mu$ is the forecast of this state from the past observations.

\paragraph{Computational cost} For large-scale applications, solving~\eqref{problem:gaussian_interpolation} by computing the closed-form expression~\eqref{eq:linear_interpolator} yields a~$\mathcal{O}(m^3 + d m)$ complexity in general, as it involves solving a~$m\times m$ linear system and computing a matrix-vector products of size~$d \times m$. In operational geophysical applications, this cost may be a bottleneck as~$d$ and~$m$ may reach prohibitively large values. To avoid such costs,~\eqref{problem:gaussian_interpolation} is solved by such as conjugate gradient~\citep{fletcher1964function}. In the data assimilation community, this variational approach for the estimation of a large-scale geophysical spatial state is called~3D-Var~\citep{courtier1998ecmwf}.

\subsection{Spatio-temporal data assimilation}
So far, the prior knowledge of the state has taken the form of a Gaussian distribution, which can capture the proximity of the searched state to an estimate, and the correlations of one state variable to another.
 Least squares interpolation then searches for the state most faithful to the data, within a fluctuation zone around the estimate. Although simple and analytically solvable, this approach does not use signal physics equations as prior information.

In the 1990s, the quality of data assimilation analyses improved significantly by incorporating a physical model to the reconstruction prior, leading to the state-of-the-art variational assimilation algorithm 4D-Var~\citep{le1986variational}. This algorithm is a generalization of~3D-Var to time-distributed observations, where the estimated signal~$x$ is a temporal sequence  of the spatial geophysical state on a time window, \textit{i.e.} a trajectory, rather than one single snapshot. The temporal dimension allows formulating the system's dynamical equations as a constraint for the signal. The reconstruction algorithm is applied sequentially on a sliding time window, in combination with a forecasting model, to produce regularly updated estimates of the meteorological variables. Alongside~4D-Var, other algorithms exist for data assimilation of dynamical systems, including sequential methods such as the celebrated~Kalman filter, and its extensions to nonlinear models~\citep{jazwinski2007stochastic}, and to ensembling~\citep{evensen2003ensemble}. In this work, we focus on the so-called weak-constraint 4D-Var algorithm~\citep{tremolet2007model, fisher2012weak}, which we briefly explain next. Weak-constraint 4D-Var has the advantage being naturally related to the~Bayesian formulation~\eqref{problem:MAP_interpolation}, and is used in operational systems.

For simplicity, we abstract from the time dimension in our mathematical formalism, and still denote the spatio-temporal signal as~$x \in \R^d$. The knowledge of a physical dynamical model materializes as knowledge of a prior distribution~$U(x)$ in~\eqref{problem:MAP_interpolation}, which can be computed and differentiated through with respect to~$x$. In geophysics, this model is typically a fluid dynamics simulator, and its gradients are computed using the adjoint method~\citep{talagrand1987variational}. Hence, the resulting~$U(x)$ is more complex and more informative than a~Gaussian prior, but comes with heavy computational costs. 
In the remained of this work, we assume that the observational processes are linear:~$h(x) = H x$. In practice,~$h$ is nonlinear and is sequentially approximated by its linear approximation. We argue that linearizing the physical model is computationally far more expensive than linearizing the observational process, and hence that considering only linear observations does not severely restrict the problem generality.

The weak-constraint 4D-Var algorithm aims at minimizing~\eqref{problem:MAP_interpolation} by a Gauss-Newton descent algorithm~\citep{gauss1877theoria}, with line-serach correction~\citep{nocedal1999numerical}. More precisely, a sequence of estimates~${\{z_k , 1 \leq k \leq \ell \}}$  approximating the reconstruction signal is iteratively computed. At each iteration~$k$, the objective function is approximated by its quadratic expansion in the vicinity of~$z_k \in \R^d$. Specifically, the prior term is approximated as
\begin{equation}
    \label{eq:prior_expansion}
\begin{aligned}
        U(x) \simeq U(z) &+ \nabla U(z)\transp{} {(x-z)}
        \\
        &+ \frac{1}{2}\transp{(x-z)} \nabla^2 U(z) (x-z).
\end{aligned}
\end{equation}
We may express expansion~\eqref{eq:prior_expansion} as a~Gaussian log-likelihood:
\begin{equation}
    \begin{aligned}
        \label{eq:local-prior}
        U(x) \simeq \frac{1}{2}\Vert x- \mu(z) \Vert^2_{P(z)^{-1}},
    \end{aligned}
\end{equation}
with
\begin{subequations}
    \begin{equation}
        P(z)  \simeq \nabla^2 U(z)^{-1},
    \end{equation}
    \begin{equation}
        \mu(z) = z - P(z)^{-1}\nabla U (z) ,
    \end{equation}
\end{subequations}
 the approximation above referring to the gradient-Hessian approximation. 


Weak-constraint 4D-Var is described in~Algorithm~\ref{algorithm:weak-4Dvar}. We see that the sequence of estimates~$(z_k)$ is iterated with a recursion of the form
    \begin{equation}
\begin{aligned}
            &x_{k} = A(z_k, y)
            \\
            &z_{k+1} = z_k + \alpha_k ({x}_{k} - z_k),
\end{aligned}
    \end{equation}
where assimilation operator~$A$ improves the current estimate~$z$ using the observations and the local approximation of the model, by performing a local optimal interpolation:
\begin{equation}
    \label{eq:4Dvar_operator}
    A(z, y) =x_{ \mathrm{MAP}} (y; \mu(z), P(z)).
\end{equation}

\paragraph{Limitations} The 4D-Var algorithm represents the state of the art for data assimilation in geophysics, and is deployed in operational meteorological centers. Its main limitation is the high computational cost of simulating and differentiating through the physical model. In Algorithm~\ref{algorithm:weak-4Dvar}, each computation of~$P_k$ and~$\mu_k$ comes with a large cost in addition to the cost of computing~\eqref{eq:4Dvar_operator}, hence limiting the method's accuracy. Note that this method may also be viewed as an application of the iterative Kalman smoother~\citep{250476, menard1996application, fisher2005equivalence, mandel20134dvar}. As is well known, an additional limitation of this method is that the non-convexity of~$U$ may lead to a complex minimization landscape, making the descent algorithm likely to be stuck in local minima~\citep{gratton2007approximate,mandel20134dvar}.
In the next section, we propose to overcome these limitations by learning operator~$A$ from data.

\section{Neural data assimilation}
\label{section:assimilation_neural}

Deep neural networks hold great promise for solving inverse problems \citep{bai2020deep}, as they can help recover the corrupted signal by using the large amount statistical information acquired on a training dataset. For the data assimilation problem in meteorology or oceanography, the ground truth signals~$x$ are not available as the geophysical systems are not observed. However, a promising research direction consists in training a deep neural network to learn a prior on high-resolution simulations, or on the reanalysis datasets such as ERA5, like neural weather models~\citep{ben2024rise}.

Deep learning approaches to inverse problems may be separated in two categories~\citep{mukherjee2021end}. A first category of algorithms aims at learning a prior~$U(x)$ from a training dataset, using a neural network, independently of the inverse problem. Once trained, the learned prior can be adapted to a reconstruction algorithm to reconstruct the signal. These algorithms are often called~``plug-and-play", as the trained neural prior can be used for any downstream inverse problem. In a second category of algorithms, referred to as~``end-to-end'' learning algorithms,  the neural network is explicitly trained to solve the inverse problem. In this case, the training consists of minimizing the neural network's reconstruction error, based on a dataset of state and observations pairs~$(x^{(i)},y^{(i)})$.

One challenge in training end-to-end algorithms is the multiplicity of possible observation processes: the trained neural network must be compatible with all possible~$(x, y)$, and hence with varying observation processes~$H$, with different dimensions~$m$ for the observations. It should therefore model only the prior distribution~$U(x)$, and not depend directly on the observation process~$H$.

\subsection{Neural assimilation operator}

\begin{algorithm}[t]
    \caption{Incremental weak-constraint 4D-Var}
    \label{algorithm:weak-4Dvar}
    \begin{algorithmic}
        \STATE \textbf{input} observation vector~$y \in \R^m$, observation matrix~${H}$,~iteration number~$\ell$, initial estimate~$z_0$, tangent linear physical model~$\mu$,~$P$
        \STATE \textbf{output} state estimation $z_\ell$
        \STATE \textbf{initialize} $z_0 := x_0$
        \FOR{$0 \leq k \leq \ell-1$}
        \STATE compute $P_k :=P(z_k)$, \; $\mu_k = \mu(z_k)$
        \STATE  estimate ${x}_{k} =  \mathrm{MAP} (y; \mu_k, P_k)$
        \STATE compute line search parameter~$\alpha_k$
        \STATE  update $z_{k+1} = z_k + \alpha_k ({x}_{k} - z_k)$
        \ENDFOR
    \end{algorithmic}
\end{algorithm}

We adopt an end-to-end learning approach, and we aim at learning a neural assimilation algorithm by minimizing a reconstruction error. We observe that, unlike other inverse problems such as image inpainting, data assimilation often starts with a first physically plausible estimate~$z$ of the unknown state. Therefore, rather than learning to interpolate the observations from scratch, we train a neural network to improve the state estimate given~$z$. Drawing inspiration from the 4D-Var algorithm, we learn an assimilation operator~$A(z, y; \theta)$, where~$\theta$ denotes the parameter vector of a neural network. As in~\eqref{eq:local-prior}, we model the local prior distribution conditioned on~$z$ as a Gaussian prior
\begin{equation}
    \label{eq:neural_prior}
    x | z\sim \mathcal{N}(\mu(z; \theta), P(z; \theta)),
\end{equation}
where $\mu(z; \theta)$ and $P(z; \theta)$ are trainable neural networks. Given this Gaussian prior, the observations are incorporated by solving the least-squares interpolation~\eqref{problem:gaussian_interpolation}:
\begin{equation}
    \label{eq:neural_assimilation-operator}
    A(z, y ; \theta) =x_{ \mathrm{MAP}} (y; \mu(z; \theta), P(z;\theta)).
\end{equation}

\begin{algorithm}[t]
    \caption{Incremental neural data assimilation}
    \label{algorithm:neural-4Dvar}
    \begin{algorithmic}
        \STATE \textbf{input} observation vector~$y \in \R^m$, observation matrix~${H}$,~iteration number~$\ell$, initial estimate~$z_0$, neural models~$\mu$,~$P$, trained parameter~$\theta$
        \STATE \textbf{output} state estimation $z_\ell$
        \STATE \textbf{initialize} $z_0 := x_0$
        \FOR{$0 \leq k \leq \ell-1$}
        \STATE compute $P_k :=P(z_k; \theta, s_k)$, \; $\mu_k = \mu(z_k; \theta, s_k)$
        \STATE  estimate ${x}_{k} =  \mathrm{MAP} (y; \mu_k, P_k)$
        \STATE  compute temperature parameter~$s_k$
        \STATE  update $z_{k+1} = z_k + s_k({x}_{k} - z_0)$ 
        \ENDFOR
    \end{algorithmic}
\end{algorithm}

\paragraph{Versatility} As we pointed out, the trained neural network should be compatible with arbitrary observation processes. 
 By formulating it as the solution of a~$y$-dependent interpolation problem, our assimilation operator~\eqref{eq:neural_assimilation-operator} is defined for any observation process~$(H, y)$, although the underlying neural networks models only the  prior distribution. In particular, the neural networks involved depend neither on~$y$, nor on~$H$: the neural assimilation operator~\eqref{eq:neural_assimilation-operator} combines the observations with a neural prior~\eqref{eq:neural_prior} of the state through the computation of a maximum likelihood estimator, and this computation is valid for any~$(H, y)$ pair for the same neural network. At prediction time, the trained neural networks~$\mu(z;\theta), P(z;\theta)$ may be used to assimilate a new observation~$y$ obtained from an arbitrary observation process~$H$ by solving~\eqref{eq:neural_assimilation-operator}.
 
 \paragraph{Training} 
 Given a dataset~$(x^{(i)}, y^{(i)},  z_0^{(i)})$ consisting of signals~$x^{(i)}$ and partial observations~$y^{(i)}$ obtained from different observation processes~$H^{(i)}$, supplemented with coarse estimates~$z^{(i)}$ of the signal, the neural prior~\eqref{eq:neural_prior} is trained to minimize the reconstruction error with the following objective:
\begin{equation}
    \label{eq:training_objective}
    \begin{aligned}
        \underset{\theta \in \R^n}{\mathrm{minimize}} \quad     
        & \sum\limits_{i=1}^N 
         \Vert A(z^{(i)}, y^{(i)}; \theta) - x^{(i)} \Vert^2
        \\
        \text{with}  \quad 
        &A(z, y; \theta) =x_{ \mathrm{MAP}} (y; \mu(z; \theta), P(z;\theta)).
    \end{aligned}
\end{equation}
We train our model by minimizing~\eqref{eq:training_objective} using stochastic gradient descent, with the~ADAM optimizer~\citep{2015-kingma}.
This training objective takes the form of a bi-level optimization problem.
Solving the inner optimization problem involves computing the optimal interpolation~\eqref{eq:linear_interpolator}, which is computed solving a linear system of size~$m$. We need to propagate the gradients with respect to~$\theta$ through this no-trivial operation during training. This may be handled by implicit differentiation, allowing to compute the gradients of the solution with respect to~$\theta$, without explicitly inverting the system's matrices~\citep{johnson2012notes}.

This training objective is similar to that of~\citep{fablet2021learning}, where a neural interpolator called~4DVarNet is used to learn both the global prior~$U(x)$ and the minimization algorithm of~\eqref{problem:MAP_interpolation}, rather than a local operator~$A(z, y) \mapsto x$. In our case, however, the inner optimization problem~\eqref{problem:MAP_interpolation} can be solved explicitly because the cost is quadratic. In contrast, it is only approximately solved in the case of 4DVarNet, due to the non-convexity of the inner cost. 

\paragraph{Computational cost} As we mentioned, the large size of the targeted physical systems requires carefully considering the computational cost of the data assimilation methods. 
We model~$P$ as a band matrix, hence limiting both the memory storage to a~$\mathcal{O}(d)$ cost and the computational complexity of solving the linear system in~\eqref{eq:linear_interpolator} using the Thomas algorithm~\citep{datta2010numerical}. This structure also imposes a temporal structure in the signal.

\subsection{Incremental neural data assimilation}

Since our assimilation operator is trained to reconstruct the signal from a coarse approximation, a one-shot reconstruction is likely to yield blurry results. To improve reconstruction, we may iterate this operator, with the aim of progressively improving the reconstruction signal. Building on the recent advances of cold diffusion~\citep{bansal2024cold}, we propose an iterative strategy aiming at reconstructing the signal in a coarse-to-fine fashion. We introduce a scalar temperature parameter~$0 \leq s \leq 1$ modeling the coarseness of the reconstruction, and we allow our neural prior to depend on~$s$ as~$\mu(z;\theta, s)$,~$P(z;\theta, s)$. Intuitively, the prior should be coarser  for larger values of~$s$, and become sharper and more local as~$s \rightarrow 0$.  We provide estimates~$z_{k}^{}$  at different temperature levels~$\{ s_1 \geq \dots \geq s_\ell \}$ as linear interpolations
~between~$z_{0}^{}$ and~$z_{\ell}^{}:=x^{}$:

\begin{equation}
    \label{eq:interpolations}
    z_{k}^{(i)} = s_k z_{0}^{(i)} + (1- s_k)z_{\ell}^{(i)}.
\end{equation}
Our training objective is adapted as
\begin{equation}
    \label{eq:temperature_objective}
    \underset{\theta \in \R^n}{\mathrm{minimize}} \quad     
     \sum\limits_{k=1}^\ell
     \sum\limits_{i=1}^N 
      \Vert A(z_{k}^{(i)}, y^{(i)}; \theta, s_k) - x^{(i)} \Vert^2.
\end{equation}
At prediction time, the signal is reconstructed by iteratively applying~$A(z,y ;\theta, s)$ following the sampling algorithm introduced in~\citep{bansal2024cold}. We provide a detailed description of our iterative reconstruction method in~Algorithm~\ref{algorithm:neural-4Dvar}. 



\section{Experiments on physical systems}
\label{section:assimilation_experiments}

In order to evaluate the performances of our data assimilation algorithm, we experiment on two simulated dynamical systems: the pendulum and the Lorenz 63 dynamical systems. We train our neural model on a dataset generated from the dynamical system with different trajectories~$x$ sampled from random initial conditions, and different observation processes, leading to various~$(x, y)$ pairs for the same~$x$. Our JAX implementation of our neural assimilation algorithm is available online at~{\footnotesize \url{https://github.com/MB-29/assimilation}}.

\paragraph{Architecture} We take for~$\mu(z; \theta, s)$ and~$P(z;\theta, s)$ two fully-connected neural networks of depth 4 and width 32. The dependence with respect to~$s$ is implemented as a positional embedding. The~$d \times d$ matrix~$P$ is modeled as a band matrix with bandwidth~$b = 2\varphi$, with~$\varphi$ the phase space dimension. 

\paragraph{Baselines} We compare our neural assimilation algorithm with various baseline. Each method starts from a first guess estimate~$z_0$ of the signal, computed by performing a~Gaussian interpolation from the observations (see below). We implement the weak-constraint 4D-Var algorithm as a Levenberg-Marquardt Gauss-Newton Algorithm using the~JAXopt implementation~\citep{jaxopt_implicit_diff}, and the~Diffrax library for differentiating through differential equation solvers~\citep{kidger2021on}.
As an ablation, an ``unconditional'' cold diffusion model is trained to restore the signal by minimizing objective~\eqref{eq:temperature_objective} without the information provided by the observations. It is then applied following~Algorithm~\ref{algorithm:neural-4Dvar} just as our neural assimilation algorithm, without using~$y$. The resulting reconstructed signal depends on the observations only through the first estimate~$z_0$, which is computed to match~$y$, but the neural network is trained to compute the next iterates by increasing only the prior term~$U(x)$ in~\eqref{problem:MAP_interpolation}, not the observation likelihood.

\subsection{Pendulum}

We start with the pendulum, which is arguably one of the simplest nonlinear physical systems. 
Importantly, the pendulum is simple enough to be decently approximated by linear dynamics. It can be shown that a linear dynamical model with Gaussian model noise yields a~Gaussian prior distribution for the trajectory~$x$. Therefore, a natural first guess for the pendulum consists in the quadratic least-squares estimator~$z_0 := x_{\mathrm{MAP}}(y; \mu_0, P_0)$, where~$\mu_0$ and~$P_0$ can be computed analytically as a function of the initial condition distribution and the pendulum's linear model. Starting from this estimate, we run the baselines and our neural assimilation algorithm.

\paragraph{Data} We generate discrete trajectories~$x^{(i)}$ of~$T=100$ time steps from the nonlinear pendulum dynamics with random initial conditions sampled in phase space, which is of dimension~$2$, hence~$d = 2\times 100=200$. The observations are generated by observing the pendulum's position at sparse time steps, with~Gaussian observation noise~$\xi \sim \mathcal{N}(0, \rho^2 I_m)$, with~$\rho = 0.01$. 

\paragraph{Experimental setup} We train an adaptation operator to reconstruct the signal in one shot from~$z_0$, following~\eqref{eq:training_objective}. At prediction time, we apply the trained neural assimilation map~$A(z;y; \theta)$ to~$z_0$ on a separate independent dataset.

\begin{figure}[ht]
    \begin{center}
        \includegraphics[width=\columnwidth]{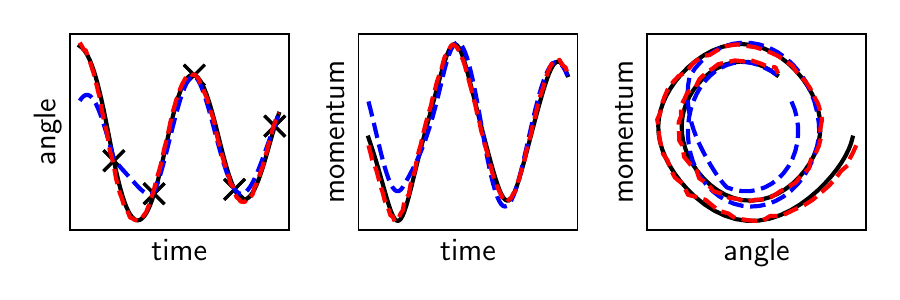}
        \par
        \includegraphics[height=.9cm]{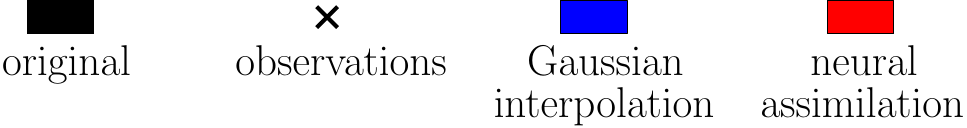}
        \caption{Reconstructed trajectories for the pendulum.}
        \label{fig:pendulum}
    \end{center}
\end{figure}

\paragraph{Results} Reconstruction samples are presented in~Figure~\ref{fig:pendulum}. While the linear model fails at reconstructing the trajectories outside the linearization zone (angle and momentum close to 0), one application of our neural assimilation operator accurately reconstructs the signal. The performances of the various methods are shown in~Table~\ref{table:assimilation_results}. Although the pendulum is simple enough for all the methods to accurately reconstruct the signal, we see that the computational gain offered by a train neural network is considerable with respect to computing the physical model.

\subsection{Lorenz 63}

We now turn to a more complex system.
The Lorenz system is a simplified physical model for for atmospheric convection~\citep{lorenz1963deterministic}. Three variables are governed by the following set of coupled nonlinear ordinary differential equations:
\begin{equation}
    \label{eq:lorenz}
    \begin{aligned}
        \frac{\ud u_1}{\ud t} &= \sigma(u_2 - u_1)
        \\
        \frac{\ud u_2}{\ud t} &= \rho u_1 -u_2 - u_1 u_3
        \\
        \frac{\ud u_3}{\ud t} &= u_1 u_2 - \beta u_3.
    \end{aligned}
\end{equation}
We set $\sigma = 10$, $\rho = 28$ and $\beta = 8/3$, values for which the system is known to exhibit chaotic solutions. We sample the initial conditions in the system's stationary distribution, following the experimental setup of~\citep{NEURIPS2023_7f7fa581}. 

\begin{figure}[ht]
    \begin{center}
        \includegraphics[width=\columnwidth]{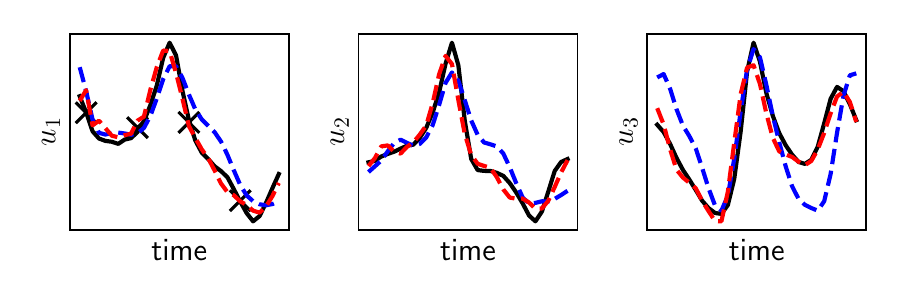}
        \par
        \vspace{-0.3cm}
        \includegraphics[width=.4\columnwidth]{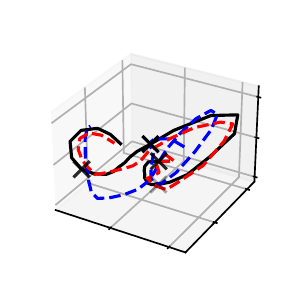} \quad 
        \raisebox{.5\height}{\includegraphics[width=.35\columnwidth]{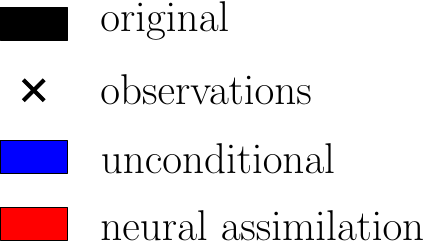}}
        \par
        \caption{Reconstructed trajectories for the Lorenz 63 system.}
        \label{fig:condition}
    \end{center}
\end{figure}

\begin{figure}[ht]
    \begin{center}
    \includegraphics[width=\columnwidth]{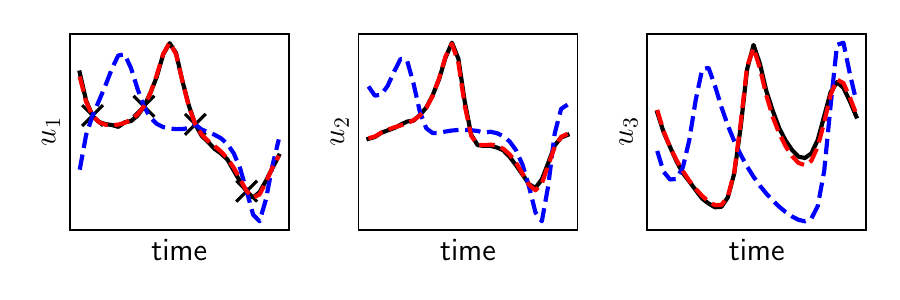}
    \par
    \vspace{-0.3cm}

    \includegraphics[width=.4\columnwidth]{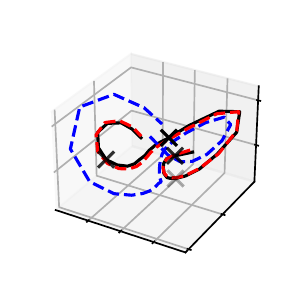}   
        \raisebox{.5\height}{\includegraphics[width=.35\columnwidth]{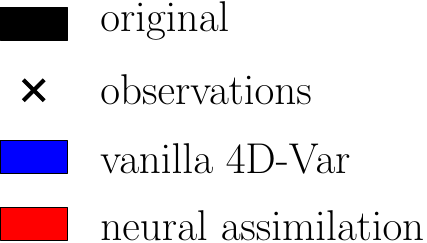}}
        \quad 
    \par
    \caption{Output of 4D-Var from various initializations.}
    \label{fig:correction}
    \end{center}
    \end{figure}

    \paragraph{Data} We generate datasets of trajectories by integrating~\eqref{eq:lorenz} between time steps of length~${\ud t = 0.025}$, and adding a small amount of~Gaussian noise~$\eta \sim \mathcal{N}(0, \ud t I_3)$ at each time step. The number of time steps is~$T = 32$, hence~$d=96$.
    We normalize each component of the trajectory to have zero mean and unit variance.
    The observations are sparse samples from the first component~$u_1$ only, with observation noise of size 0.05.
    We take for the initial state estimate~$z^{(i)}$ the maximum likelihood interpolation of~$y^{(i)}$ under the moment-matching Gaussian distribution of~$x^{(i)}$, which is the coarse Gaussian approximate of~$p(x)$. More precisely,~$z_0^{(i)} = x_{\mathrm{MAP}}(y^{(i)}; \hat{\mu}, \hat{P})$, with~$\hat{\mu}$ and~$\hat{P}$ the empirical mean and the empirical covariance of~$\{x^{(i)}\}$. We define~$\{z_k^{(i)}\}$ as in~\eqref{eq:interpolations} with regular spacing~$s_k  =1- k/(\ell+1)$. We take~$\ell = 5$.

    \paragraph{Experimental setup} We train our neural assimilation operator to reconstruct the signal at different temperatures following~\eqref{eq:temperature_objective}. At prediction time, we apply~Algorithm~\ref{algorithm:neural-4Dvar} for the neural methods, along with the~4D-Var algorithm~(Algorithm~\ref{algorithm:weak-4Dvar}). Furthermore, in order to establish a link between our new neural method and traditional assimilation methods, we investigate how the output of the neural method, which is a priori  not interpretable, may be transformed into a plausible physical signal. To do this, we correct these estimates with several iterations of 4D-Var on top of the  neural estimate of the signal, until the objective function~\eqref{problem:MAP_interpolation} becomes lower than 0.05. As a result, the new output is constrained to satisfy the physical model, but potentially at a lower cost than if we had started from scratch because the initialization that we provided is already close to the true signal.

    \begin{table}
        \caption{Performances of the various approaches. The computational time unit is the run-time of the fastest of the algorithms at prediction time.}
        \label{table:assimilation_results}
        \vskip 0.15in
        \begin{center}
        \begin{small}
        \resizebox{.5\textwidth}{!}{%
        \begin{tabular}{lcccr}
        \hline
        Method & 4D-Var & Cold diffusion & Neural assimilation\\
        \hline
        Pendulum  error    & 0.1& 0.15& 0.12 \\
        \hline
        Lorenz 63  error    & 0.9& 1.1& 0.5 \\
        \hline
        Computational time    & 10& 1& 2 \\
        \hline
        \end{tabular}}
        \end{small}
        \end{center}
        \vskip -0.1in
        \end{table}
    
    \paragraph{Results} Figure~\ref{fig:condition} shows reconstruction samples from the baselines and from our method, and~Table~\ref{table:assimilation_results} shows the average reconstruction error for the various methods. We can see that our neural data assimilation algorithm can reconstruct the signal while staying close to the observations. In contrast, the unconditional baseline cannot efficiently improve both the signal likelihood and the data fidelity. Compared to~4D-Var, our neural approach offers considerable computational gains, and good accuracy in these experiments. Further, we compare the reconstructed signals corrected by 4D-Var for an observation sample in~Figure~\ref{fig:correction}, where fixed number of 4D-Var iterations are applied to two different initializations: the Gaussian first-guess and the neural reconstruction of our algorithm. The initialization provided by our method allows to recover the original signal with very high accuracy by running few steps of 4D-Var on top of the neural estimate, while the 4D-Var algorithm  with Gaussian initialization (``vanilla'') leads to an inaccurate local minimum. Importantly, the improvement with respect to a Gaussian initialization is significant, both in terms of reconstruction error and in terms of number of iterations, as the 4D-var algorithm converged after 4 iterations from the neural initialization and 23 iterations from the~Gaussian initialization. We further discuss the comparison between deep learning data assimilation approaches and 4D-Var in~Section~\ref{section:assimilation_conclusion}.

    \section{Related work}
    \label{section:assimilation_related}
    
    The state of the art methods for data assimilation are the 4D-Var algorithm~\citep{le1986variational,tremolet2007model} and the ensemble Kalman filter~\citep{evensen2003ensemble,bocquet2014introduction}. The statistical component in these approaches lies in the definition of covariance matrices for the background state estimates, for the model and for the observations. The numerical cost of computing the physical model and its linear tangent local approximation may be considerable for large systems.  
    
    In recent years, several deep learning algorithms have been proposed for the data assimilation problem. Building on diffusion models~\citep{ho2020denoising}, \cite{NEURIPS2023_7f7fa581} propose a data assimilation method based on score-based diffusion. This approach proceeds in a plug-and-play fashion, and sampling from the posterior distribution relies on an approximation that is computed on the trained model. Among ``end-to-end'' deep learning approaches for data assimilation, the one that is closest related to ours is the 4DVarNet algorithm of~\cite{fablet2021learning}, which aims to directly train a neural network to minimize the reconstruction error. The complex prior modeled by the neural network is non-Gaussian, and estimating the maximum a posteriori reconstruction in this framework relies on non-convex optimization.

    \section{Conclusion}
    \label{section:assimilation_conclusion}
    
    In this work, we have shown how deep learning methods may be applied to the data assimilation problem. Our neural method models in a coarse-to-fine fashion and is trained to minimize the reconstruction error. Importantly, we have shown how such a deep learning method may be used in combination with a traditional data assimilation method to enhance the reconstruction accuracy and reduce the computational time.
    
    We believe that deep learning methods alone might not be accurate enough to completely outperform traditional physics-based approaches such as 4D-Var. While our neural approach had good reconstruction results on the presented simulated physical systems, it should be noted that the small size of these systems allows for the neural network to learn the stationary distribution from a reasonably small dataset. For real-life systems, it is unlikely that a neural network can accurately generalize the learned signal outside a training dataset, where the physics may be complex and fairly different from what the model has seen. In contrast, physics-based approaches  are far more general, as the simulated physical laws are accurate everywhere in the state space. Therefore, using a deep learning algorithm to provide an approximate solution, and using it as an input to 4D-Var to reduce the number of iterations seems like a good trade-off benefitting the best of both words.
    
    In future work, it would be interesting to apply our method to physical systems of larger scale, and to explore how the computational burden of data assimilation may be further reduced on such high-dimensional systems. Another important aspect that is crucial for data assimilation is uncertainty quantification, for which there has been recent progress in the deep learning community~\citep{arcucci2021deep, corso2022diffdock}.

\clearpage
\bibliography{references}

\begin{thebibliography}{39}
\providecommand{\natexlab}[1]{#1}
\providecommand{\url}[1]{\texttt{#1}}
\expandafter\ifx\csname urlstyle\endcsname\relax
  \providecommand{\doi}[1]{doi: #1}\else
  \providecommand{\doi}{doi: \begingroup \urlstyle{rm}\Url}\fi

\bibitem[Arcucci et~al.(2021)Arcucci, Zhu, Hu, and Guo]{arcucci2021deep}
Arcucci, R., Zhu, J., Hu, S., and Guo, Y.-K.
\newblock Deep data assimilation: integrating deep learning with data
  assimilation.
\newblock \emph{Applied Sciences}, 11\penalty0 (3):\penalty0 1114, 2021.

\bibitem[Bai et~al.(2020)Bai, Chen, Chen, and Guo]{bai2020deep}
Bai, Y., Chen, W., Chen, J., and Guo, W.
\newblock Deep learning methods for solving linear inverse problems: Research
  directions and paradigms.
\newblock \emph{Signal Processing}, 177:\penalty0 107729, 2020.

\bibitem[Bansal et~al.(2024)Bansal, Borgnia, Chu, Li, Kazemi, Huang, Goldblum,
  Geiping, and Goldstein]{bansal2024cold}
Bansal, A., Borgnia, E., Chu, H.-M., Li, J., Kazemi, H., Huang, F., Goldblum,
  M., Geiping, J., and Goldstein, T.
\newblock Cold diffusion: Inverting arbitrary image transforms without noise.
\newblock \emph{Advances in Neural Information Processing Systems}, 36, 2024.

\bibitem[Bell \& Cathey(1993)Bell and Cathey]{250476}
Bell, B. and Cathey, F.
\newblock The iterated kalman filter update as a gauss-newton method.
\newblock \emph{IEEE Transactions on Automatic Control}, 38\penalty0
  (2):\penalty0 294--297, 1993.
\newblock \doi{10.1109/9.250476}.

\bibitem[Ben~Bouall{\`e}gue et~al.(2024)Ben~Bouall{\`e}gue, Clare, Magnusson,
  Gascon, Maier-Gerber, Janou{\v{s}}ek, Rodwell, Pinault, Dramsch, Lang,
  et~al.]{ben2024rise}
Ben~Bouall{\`e}gue, Z., Clare, M.~C., Magnusson, L., Gascon, E., Maier-Gerber,
  M., Janou{\v{s}}ek, M., Rodwell, M., Pinault, F., Dramsch, J.~S., Lang,
  S.~T., et~al.
\newblock The rise of data-driven weather forecasting: A first statistical
  assessment of machine learning-based weather forecasts in an operational-like
  context.
\newblock \emph{Bulletin of the American Meteorological Society}, 2024.

\bibitem[Blondel et~al.(2021)Blondel, Berthet, Cuturi, Frostig, Hoyer,
  Llinares-L{\'o}pez, Pedregosa, and Vert]{jaxopt_implicit_diff}
Blondel, M., Berthet, Q., Cuturi, M., Frostig, R., Hoyer, S.,
  Llinares-L{\'o}pez, F., Pedregosa, F., and Vert, J.-P.
\newblock Efficient and modular implicit differentiation.
\newblock \emph{arXiv preprint arXiv:2105.15183}, 2021.

\bibitem[Bocquet et~al.(2014)]{bocquet2014introduction}
Bocquet, M. et~al.
\newblock Introduction to the principles and methods of data assimilation in
  geosciences.
\newblock \emph{Notes de cours, {\'E}cole des Ponts ParisTech}, 2014.

\bibitem[Bouttier \& Courtier(2002)Bouttier and Courtier]{bouttier2002data}
Bouttier, F. and Courtier, P.
\newblock Data assimilation concepts and methods march 1999.
\newblock \emph{Meteorological training course lecture series. ECMWF},
  718:\penalty0 59, 2002.

\bibitem[Corso et~al.(2022)Corso, St{\"a}rk, Jing, Barzilay, and
  Jaakkola]{corso2022diffdock}
Corso, G., St{\"a}rk, H., Jing, B., Barzilay, R., and Jaakkola, T.
\newblock Diffdock: Diffusion steps, twists, and turns for molecular docking.
\newblock \emph{arXiv preprint arXiv:2210.01776}, 2022.

\bibitem[Courtier et~al.(1998)Courtier, Andersson, Heckley, Vasiljevic, Hamrud,
  Hollingsworth, Rabier, Fisher, and Pailleux]{courtier1998ecmwf}
Courtier, P., Andersson, E., Heckley, W., Vasiljevic, D., Hamrud, M.,
  Hollingsworth, A., Rabier, F., Fisher, M., and Pailleux, J.
\newblock The ecmwf implementation of three-dimensional variational
  assimilation (3d-var). i: Formulation.
\newblock \emph{Quarterly Journal of the Royal Meteorological Society},
  124\penalty0 (550):\penalty0 1783--1807, 1998.

\bibitem[Datta(2010)]{datta2010numerical}
Datta, B.~N.
\newblock \emph{Numerical linear algebra and applications}.
\newblock SIAM, 2010.

\bibitem[Evensen(2003)]{evensen2003ensemble}
Evensen, G.
\newblock The ensemble kalman filter: Theoretical formulation and practical
  implementation.
\newblock \emph{Ocean dynamics}, 53:\penalty0 343--367, 2003.

\bibitem[Fablet et~al.(2021)Fablet, Chapron, Drumetz, M{\'e}min, Pannekoucke,
  and Rousseau]{fablet2021learning}
Fablet, R., Chapron, B., Drumetz, L., M{\'e}min, E., Pannekoucke, O., and
  Rousseau, F.
\newblock Learning variational data assimilation models and solvers.
\newblock \emph{Journal of Advances in Modeling Earth Systems}, 13\penalty0
  (10):\penalty0 e2021MS002572, 2021.

\bibitem[Fisher et~al.(2005)Fisher, Leutbecher, and
  Kelly]{fisher2005equivalence}
Fisher, M., Leutbecher, M., and Kelly, G.
\newblock On the equivalence between kalman smoothing and weak-constraint
  four-dimensional variational data assimilation.
\newblock \emph{Quarterly Journal of the Royal Meteorological Society: A
  journal of the atmospheric sciences, applied meteorology and physical
  oceanography}, 131\penalty0 (613):\penalty0 3235--3246, 2005.

\bibitem[Fisher et~al.(2012)Fisher, Tr{\'e}molet, Auvinen, Tan, and
  Poli]{fisher2012weak}
Fisher, M., Tr{\'e}molet, Y., Auvinen, H., Tan, D., and Poli, P.
\newblock \emph{Weak-constraint and long-window 4D-Var}.
\newblock ECMWF Reading, UK, 2012.

\bibitem[Fletcher \& Reeves(1964)Fletcher and Reeves]{fletcher1964function}
Fletcher, R. and Reeves, C.~M.
\newblock Function minimization by conjugate gradients.
\newblock \emph{The computer journal}, 7\penalty0 (2):\penalty0 149--154, 1964.

\bibitem[Gauss(1877)]{gauss1877theoria}
Gauss, C.~F.
\newblock \emph{Theoria motus corporum coelestium in sectionibus conicis solem
  ambientium}, volume~7.
\newblock FA Perthes, 1877.

\bibitem[Gratton et~al.(2007)Gratton, Lawless, and
  Nichols]{gratton2007approximate}
Gratton, S., Lawless, A.~S., and Nichols, N.~K.
\newblock Approximate gauss--newton methods for nonlinear least squares
  problems.
\newblock \emph{SIAM Journal on Optimization}, 18\penalty0 (1):\penalty0
  106--132, 2007.

\bibitem[Ho et~al.(2020)Ho, Jain, and Abbeel]{ho2020denoising}
Ho, J., Jain, A., and Abbeel, P.
\newblock Denoising diffusion probabilistic models.
\newblock \emph{Advances in neural information processing systems},
  33:\penalty0 6840--6851, 2020.

\bibitem[Hoyer et~al.(2023)Hoyer, Yuval, Kochkov, Langmore, Norgaard, Mooers,
  and Brenner]{hoyer2023neural}
Hoyer, S., Yuval, J., Kochkov, D., Langmore, I., Norgaard, P., Mooers, G., and
  Brenner, M.~P.
\newblock Neural general circulation models for weather and climate.
\newblock \emph{AGU23}, 2023.

\bibitem[Jazwinski(2007)]{jazwinski2007stochastic}
Jazwinski, A.~H.
\newblock \emph{Stochastic processes and filtering theory}.
\newblock Courier Corporation, 2007.

\bibitem[Johnson(2012)]{johnson2012notes}
Johnson, S.~G.
\newblock Notes on adjoint methods for 18.335.
\newblock \emph{Introduction to Numerical Methods}, 2012.

\bibitem[Kidger(2021)]{kidger2021on}
Kidger, P.
\newblock \emph{{O}n {N}eural {D}ifferential {E}quations}.
\newblock PhD thesis, University of Oxford, 2021.

\bibitem[Kingma \& Ba(2015)Kingma and Ba]{2015-kingma}
Kingma, D.~P. and Ba, J.
\newblock Adam: A method for stochastic optimization.
\newblock In \emph{ICLR (Poster)}, 2015.
\newblock URL
  \url{http://dblp.uni-trier.de/db/conf/iclr/iclr2015.html#KingmaB14}.

\bibitem[Lam et~al.(2022)Lam, Sanchez-Gonzalez, Willson, Wirnsberger,
  Fortunato, Alet, Ravuri, Ewalds, Eaton-Rosen, Hu, et~al.]{lam2022graphcast}
Lam, R., Sanchez-Gonzalez, A., Willson, M., Wirnsberger, P., Fortunato, M.,
  Alet, F., Ravuri, S., Ewalds, T., Eaton-Rosen, Z., Hu, W., et~al.
\newblock Graphcast: Learning skillful medium-range global weather forecasting.
\newblock \emph{arXiv preprint arXiv:2212.12794}, 2022.

\bibitem[Laumont et~al.(2022)Laumont, Bortoli, Almansa, Delon, Durmus, and
  Pereyra]{laumont2022bayesian}
Laumont, R., Bortoli, V.~D., Almansa, A., Delon, J., Durmus, A., and Pereyra,
  M.
\newblock Bayesian imaging using plug \& play priors: when langevin meets
  tweedie.
\newblock \emph{SIAM Journal on Imaging Sciences}, 15\penalty0 (2):\penalty0
  701--737, 2022.

\bibitem[Le~Dimet \& Talagrand(1986)Le~Dimet and Talagrand]{le1986variational}
Le~Dimet, F.-X. and Talagrand, O.
\newblock Variational algorithms for analysis and assimilation of
  meteorological observations: theoretical aspects.
\newblock \emph{Tellus A: Dynamic Meteorology and Oceanography}, 38\penalty0
  (2):\penalty0 97--110, 1986.

\bibitem[Liang(2005)]{liang2005quantitative}
Liang, S.
\newblock \emph{Quantitative remote sensing of land surfaces}.
\newblock John Wiley \& Sons, 2005.

\bibitem[Lorenz(1963)]{lorenz1963deterministic}
Lorenz, E.~N.
\newblock Deterministic nonperiodic flow.
\newblock \emph{Journal of atmospheric sciences}, 20\penalty0 (2):\penalty0
  130--141, 1963.

\bibitem[Mandel et~al.(2013)Mandel, Bergou, and Gratton]{mandel20134dvar}
Mandel, J., Bergou, E., and Gratton, S.
\newblock 4dvar by ensemble kalman smoother.
\newblock \emph{arXiv preprint arXiv:1304.5271}, 2013.

\bibitem[M{\'e}nard \& Daley(1996)M{\'e}nard and Daley]{menard1996application}
M{\'e}nard, R. and Daley, R.
\newblock The application of kalman smoother theory to the estimation of 4dvar
  error statistics.
\newblock \emph{Tellus A}, 48\penalty0 (2):\penalty0 221--237, 1996.

\bibitem[Mukherjee et~al.(2021)Mukherjee, Carioni, {\"O}ktem, and
  Sch{\"o}nlieb]{mukherjee2021end}
Mukherjee, S., Carioni, M., {\"O}ktem, O., and Sch{\"o}nlieb, C.-B.
\newblock End-to-end reconstruction meets data-driven regularization for
  inverse problems.
\newblock \emph{Advances in Neural Information Processing Systems},
  34:\penalty0 21413--21425, 2021.

\bibitem[Mu{\~n}oz-Sabater et~al.(2021)Mu{\~n}oz-Sabater, Dutra,
  Agust{\'\i}-Panareda, Albergel, Arduini, Balsamo, Boussetta, Choulga,
  Harrigan, Hersbach, et~al.]{munoz2021era5}
Mu{\~n}oz-Sabater, J., Dutra, E., Agust{\'\i}-Panareda, A., Albergel, C.,
  Arduini, G., Balsamo, G., Boussetta, S., Choulga, M., Harrigan, S., Hersbach,
  H., et~al.
\newblock Era5-land: A state-of-the-art global reanalysis dataset for land
  applications.
\newblock \emph{Earth system science data}, 13\penalty0 (9):\penalty0
  4349--4383, 2021.

\bibitem[Nocedal \& Wright(1999)Nocedal and Wright]{nocedal1999numerical}
Nocedal, J. and Wright, S.~J.
\newblock \emph{Numerical optimization}.
\newblock Springer, 1999.

\bibitem[Pathak et~al.(2022)Pathak, Subramanian, Harrington, Raja,
  Chattopadhyay, Mardani, Kurth, Hall, Li, Azizzadenesheli,
  et~al.]{pathak2022fourcastnet}
Pathak, J., Subramanian, S., Harrington, P., Raja, S., Chattopadhyay, A.,
  Mardani, M., Kurth, T., Hall, D., Li, Z., Azizzadenesheli, K., et~al.
\newblock Fourcastnet: A global data-driven high-resolution weather model using
  adaptive fourier neural operators.
\newblock \emph{arXiv preprint arXiv:2202.11214}, 2022.

\bibitem[Rangayyan \& Krishnan(2024)Rangayyan and
  Krishnan]{rangayyan2024biomedical}
Rangayyan, R.~M. and Krishnan, S.
\newblock \emph{Biomedical signal analysis}.
\newblock John Wiley \& Sons, 2024.

\bibitem[Rozet \& Louppe(2023)Rozet and Louppe]{NEURIPS2023_7f7fa581}
Rozet, F. and Louppe, G.
\newblock Score-based data assimilation.
\newblock In Oh, A., Naumann, T., Globerson, A., Saenko, K., Hardt, M., and
  Levine, S. (eds.), \emph{Advances in Neural Information Processing Systems},
  volume~36, pp.\  40521--40541. Curran Associates, Inc., 2023.

\bibitem[Talagrand \& Courtier(1987)Talagrand and
  Courtier]{talagrand1987variational}
Talagrand, O. and Courtier, P.
\newblock Variational assimilation of meteorological observations with the
  adjoint vorticity equation. i: Theory.
\newblock \emph{Quarterly Journal of the Royal Meteorological Society},
  113\penalty0 (478):\penalty0 1311--1328, 1987.

\bibitem[Tr{\'e}molet(2007)]{tremolet2007model}
Tr{\'e}molet, Y.
\newblock Model-error estimation in 4d-var.
\newblock \emph{Quarterly Journal of the Royal Meteorological Society: A
  journal of the atmospheric sciences, applied meteorology and physical
  oceanography}, 133\penalty0 (626):\penalty0 1267--1280, 2007.

\end{thebibliography}
\bibliographystyle{icml2024}





\end{document}